\documentclass[11pt]{article}

\usepackage{acl}
\usepackage{amsmath}
\usepackage{times}
\usepackage{latexsym}

\usepackage{booktabs}
\usepackage{amssymb} 

\usepackage[T1]{fontenc}

\usepackage[utf8]{inputenc}

\usepackage{microtype}

\usepackage{inconsolata}

\usepackage{graphicx}

\title{Pose-Based Sign Language Spotting via an End-to-End Encoder Architecture}

\author{
  Samuel Ebimobowei Johnny\textsuperscript{1} \quad
  Blessed Guda\textsuperscript{1,2} \\
  Emmanuel Enejo Aaron\textsuperscript{1} \quad
  Assane Gueye\textsuperscript{1,2} \\
  \texttt{\{sjohnny, blessedg, eaaron, assaneg\}@andrew.cmu.edu} \\
  \textsuperscript{1}Carnegie Mellon University Africa, Kigali, Rwanda \\
  \textsuperscript{2}Carnegie Mellon University, Pittsburgh, USA
}

\begin{document}
\maketitle
\begin{abstract}
Automatic Sign Language Recognition (ASLR) has emerged as a vital field for bridging the gap between deaf and hearing communities. However, the problem of sign-to-sign retrieval or detecting a specific sign within a sequence of continuous signs remains largely unexplored. We define this novel task as Sign Language Spotting. In this paper, we present a first step toward sign language retrieval by addressing the challenge of detecting the presence or absence of a query sign video within a sentence-level gloss or sign video.  Unlike conventional approaches that rely on intermediate gloss recognition or text-based matching, we propose an end-to-end model that directly operates on pose keypoints extracted from sign videos. Our architecture employs an encoder-only backbone with a binary classification head to determine whether the query sign appears within the target sequence. By focusing on pose representations instead of raw RGB frames, our method significantly reduces computational cost and mitigates visual noise. We evaluate our approach on the Word Presence Prediction dataset from the WSLP 2025 shared task, achieving 61.88\% accuracy and 60.00\% F1-score. These results demonstrate the effectiveness of our pose-based framework for Sign Language Spotting, establishing a strong foundation for future research in automatic sign language retrieval and verification. Code is available at \href{https://github.com/EbimoJohnny/Pose-Based-Sign-Language-Spotting}{this repository}.
 \end{abstract}

\section{Introduction}

Sign language, which globally consists of more than 300 different sign languages~\cite{UN2023SignLanguages}, was developed to address the need for effective communication for the deaf and hearing-impaired population~\cite{tunga2021pose}. Each sign language comprises a complex combination of hand gestures, facial expressions, and body movements that collectively encode the semantics and grammatical structures of spoken languages~\cite{tang2025gloss, rastgoo2024word}. However, there is still a challenge and a communication gap between the deaf and hearing community~\cite{das2024occlusion}, ~\cite{10.1016/j.eswa.2021.115601}. 
Previous works have focused on sign language translation(SLT) where researchers have attempted to translate sign language either as RGB or poses to either text(that is word word-level semantically meaningful)~\cite{yin-read-2020-better,9706762} or glosses~\cite{Zhou_2023_ICCV,LowJianHe2025SSGE}. 

Sign language recognition(SLR) could be isolated and continuous SLR. Isolated sign language (ISLR) ~\cite{electronics13071229, Baihan2024,REN2025128340} translation involves word-level focuses on recognizing individual signs 
in isolation, treating each sign as an independent classification problem. In contrast, Continuous Sign Language Recognition (CSLR)~\cite{10829616,10.1145/3742886.3756720,10204826,LowJianHe2025SSGE} involves sentence-level SL, which addresses a more challenging task of translating continuous signing sequences into semantically correct sentences or gloss annotations, requiring models to handle temporal dependencies, co-articulation effects, and 
variable-length sequences.

While significant progress has been made in recognition and translation, the ability to search, retrieve, or verify specific signs within continuous signing videos remains underexplored. This capability- known as sign spotting - is critical for applications such as SL retrieval, dictionary lookup,  and educational tools. This requires robust sign spotting capabilities, that is, the ability to locate and identify specific signs within continuous signing videos. Traditional approaches to this problem have relied on text-based intermediate representations.

For word spotting for CLSR, researchers have attempted to spot words using Large Language Models (LLMs). ~\cite{DBLP:journals/corr/abs-2308-04248} proposed using LLMs such as BERT and Word2Vec to leverage alignment to improve isolated signs from continuous signs. Their approach solves text gloss mapping using LLMs; their model provides an effective method, which was evaluated on MeieneDGS~\cite{dgscorpus_3} and BOBSL\cite{Albanie2021bobsl}.

 Recent work on sign spotting addresses the challenge of SLT by decomposing it into modular stages. Spotter+GPT~\cite{10.1145/3742886.3756720} proposes an approach to eliminate the need for SLT-specific end-to-end training, significantly reducing computational costs. Their approach extracts I3D motion and ResNeXt-101 handshape features, matches them to a sign dictionary using DTW and cosine similarity, and passes spotted signs, which are the top-k glosses, to GPT for sentence generation. While Spotter+GPT demonstrates the effectiveness of modular SLT, our work addresses a fundamentally different task: word presence verification. 

In this paper, we introduce a novel end-to-end video-to-video sign spotting framework that eliminates the need for textual or gloss-based intermediates. Given a query sign video and a sentence-level sign video, our model determines whether the query sign is present within the sentence. We adopt an encoder-only architecture with a binary classification head, operating directly on pose keypoints rather than RGB frames. This design reduces computational complexity and suppresses visual noise while maintaining discriminative spatial-temporal information. We evaluate our approach on the Word Presence Prediction dataset from the WSLP 2025 shared task\footnote{https://exploration-lab.github.io/WSLP/task/}. To the best of our knowledge, this work represents the first study to address sign language spotting purely through video-to-video matching, establishing a foundation for future research in automatic sign language retrieval, verification, and search.

\section{Methodology}

\begin{figure*}
    \centering
    \includegraphics[width=1.0\linewidth]{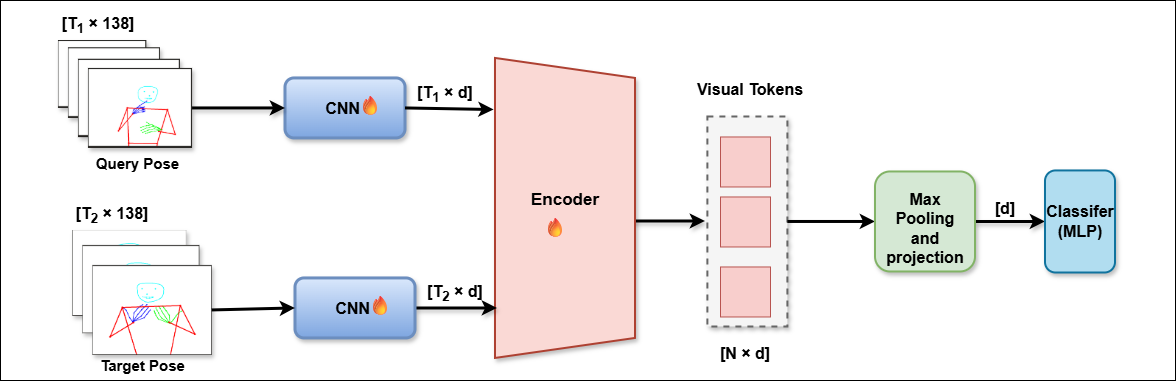}
    \caption{Architecture overview. Pose sequences are encoded using 2D CNNs and then processed by a Transformer encoder, which produces visual tokens.  The \textbf{[CLS]} token is max-pooled to predict query presence using binary cross-entropy loss ($\mathcal{L}_{BCE}$). }
    \label{fig:placeholder}
\end{figure*}

We propose a video-to-video sign spotting architecture that jointly models visual–semantic alignment and binary word presence prediction. The framework learns robust cross-modal representations that generalize across signers and sentence contexts.
Our approach consists of three main components: pose extraction, feature encoding, and presence prediction, as illustrated in Figure~\ref{fig:placeholder}.
\subsection{Pose Extraction}
\label{sub:pose}
We use MediaPipe~\cite{lugaresi2019mediapipe} to convert RGB video sequences to pose-based representations, allowing for a more generalized, efficient, and resilient architecture. For each frame, MediaPipe estimates the pose keypoints of the signer in the video. Following~\cite{johnny2025autosign}, we extract holistic pose features containing 42 hand keypoints (21 per hand), 8 body keypoints, and 19 facial landmarks. As suggested by ~\cite{johnny2025autosign}, we used only the hand and body features in this study. 

\subsection{Problem formulation}
Given a sentence sequence $\mathbf{X}_s \in \mathbb{R}^{T_s \times F}$ and a query sequence $\mathbf{X}_q \in \mathbb{R}^{T_q \times F}$, where $T_s$ and $T_q$ denote the temporal lengths of the sentence and query sequences respectively, and $F$ represents the dimensionality of the pose features, the objective is to determine whether the query sign appears within the sentence.

Let $f(\mathbf{X}_s, \mathbf{X}_q; \theta)$ be a parameterized model, where $\theta$ denotes the set of learnable parameters. The model outputs a probability score $\hat{y} = f(\mathbf{X}_s, \mathbf{X}_q; \theta)$, representing the likelihood that the query sign occurs in the given sentence. The binary prediction is made as:
The training objective is to optimize the model parameters $\theta$ by minimizing a loss function $\mathcal{L}(\theta)$ over the training data:
\[
\theta^{*} = \arg \min_{\theta} \mathcal{L}(\theta).
\]

\subsection{Pose CNN Encoder}
Each pose frame is represented as a vector $\mathbf{x}_t \in \mathbb{R}^{100}$, corresponding to 50 keypoints with 2D coordinates $(x, y)$. To preserve the spatial topology of the human skeleton, each vector is reshaped into a 2D array of size $\mathbb{R}^{50 \times 2}$, considering only the hand and body keypoints as described in Section~\ref{sub:pose}.

A 2D CNN is applied independently to each frame to extract local spatial dependencies among keypoints. Specifically, each pose frame passes through three Conv2D blocks, each consisting of a convolutional layer, Batch Normalization, and ReLU activation. These blocks progressively capture hierarchical geometric patterns while maintaining spatial coherence across keypoints.

Following the convolutional layers, an \textbf{adaptive average pooling} layer reduces the spatial dimensions to a fixed-size representation, which is then linearly projected to a feature vector of dimension $d = 128$.

This process yields a sequence of per-frame embeddings:
\begin{equation}
    \mathbf{H} = \{\mathbf{h}_t\}_{t=1}^{T} \in \mathbb{R}^{T \times 128},
\end{equation}
where $T$ denotes the total frames in the video. Each embedding $\mathbf{h}_t$ encodes the spatial structure of the signer’s body and hand poses at time step $t$. The resulting feature sequence is passed to the transformer encoder for temporal modeling.

\subsection{Visual Transformer Encoder model}
To enable temporal dependencies and cross-sequence interactions between the query and sentence embeddings, we adopt a BERT-style sequence modeling approach. Specifically, a $[\text{CLS}]$ token is prepended for global sequence-level classification, while a $[\text{SEP}]$ token is inserted to explicitly separate the \textit{query pose tokens} from the \textit{candidate pose tokens}. This design enables the model to attend across the boundary between the two sequences, allowing direct interaction between corresponding temporal segments.  

Learnable positional encodings and token-type embeddings are incorporated to preserve temporal order and to distinguish between query and candidate sequences. The transformer encoder then processes the concatenated sequence using multi-head self-attention, where the \textbf{attention scores between query and candidate pose tokens} serve as a key mechanism for measuring their semantic and spatial correspondence. These cross-sequence attention patterns help the model identify whether visual and structural similarities exist between the query sign and any segment of the candidate video, thereby assisting the sign spotting task.  



\subsubsection{Classification Loss}
Since the expected outcome is binary(present or absent), our model employs the binary cross-entropy loss(BCE) to penalize incorrect and overconfident predictions. We extract the \textbf{[CLS]} token representation via max pooling and project it to an MLP classifier to generate the corresponding logits $\hat{y} \in \mathbb{R}$. The final prediction is obtained by applying the sigmoid($\sigma$) to the logits. The binary cross-entropy loss is computed as:
\begin{equation}
\mathcal{L}_{\text{BCE}} = -\frac{1}{B}\sum_{i=1}^{B} [y_i \log p_i + (1-y_i)\log(1-p_i)]
\end{equation}

where $p = \sigma(\hat{y}_i))$, $\sigma$ is the sigmoid function , $y_i \in \{0,1\}$ is the ground truth 
label, and $B$ is the batch size. 





\section{Experiments}

\subsection{Dataset and Evaluation Metrics}
For this experiment, we evaluate our model with \textit{Word Presence Dataset}\footnote{\url{https://huggingface.co/datasets/Exploration-Lab/WSLP-AACL-2025}}, an ASL sign spotting dataset designed to determine if a query sign appears within a sentence sequence. 

The dataset comprises 25,432 sentence-query pairs constructed from 7,857 unique sentence sequences and 1,410 unique query sequences. The dataset is balanced, with equal distribution of positive (query present) and negative (query absent) samples. We employ an 80:20 train-validation split. The test set contains 1,266 unique sentence sequences and 555 unique query sequences, ensuring minimal overlap with the training distribution.  

During Evaluation, we use standard classification evaluation metrics, i.e., Accuracy, Precision, Recall, and F1-score.

\section{Implementation Details}
\label{sec:appendix}

We train the model end-to-end with an initial learning rate of 0.0005 since commonly used values (e.g., 0.001 or 0.01) resulted in suboptimal convergence, using the AdamW optimizer and a temperature of 0.07 for contrastive losses. A dropout of 0.02 was applied to prevent overfitting. Training is carried out for 50 epochs with a patience of 5  if no future improvements. This was done using a single NVIDIA L40S GPU.

To ensure our model focuses on important features, we skipped all early and late frames with no finger movement. During training, we applied different data augmentation techniques such as sequence masking, scaling, jittering, and Gaussian noise to ensure robustness.


\begin{table}[t]
\centering
\caption{Performance comparison on Word Presence Prediction (Test Set).}
\label{tab:main_results}
\footnotesize
\setlength{\tabcolsep}{6.0pt}  
\renewcommand{\arraystretch}{1.2} 
\begin{tabular}{lcccc}
\toprule
\textbf{Method} & \textbf{Acc.} & \textbf{F1} & \textbf{Prec.} & \textbf{Rec.} \\
\midrule
Ours (1D CNN) & 60.95 & \textbf{59.62} & 62.70 & 61.01 \\
\textbf{Ours (2D CNN)} & \textbf{61.66} & 58.42 & \textbf{67.16} & \textbf{61.74} \\
\bottomrule
\end{tabular}
\end{table}

\subsection{Evaluation on Test Set}





Table~\ref{tab:main_results} presents our results on the test set. Given that this is a novel task introduced in the WSLP 2025 shared task, with no prior work to the best of our knowledge, hence no baseline to compare against. Our 2D-CNN approach achieves 61.66\% 
accuracy, outperforming linear(1D-CNN) projection. Notably, 2D-CNN significantly improves 
precision, indicating fewer false positives, though F1 
slightly decreases due to the precision-recall trade-off.

\begin{table}[t]
\centering
\caption{Ablation study on validation set.}
\label{tab:ablation}
\footnotesize
\setlength{\tabcolsep}{6.0pt}
\renewcommand{\arraystretch}{1.1} 
\begin{tabular}{lcccc}
\toprule
\textbf{Configuration} & \textbf{Acc.} & \textbf{F1} & \textbf{Prec.} & \textbf{Rec.} \\
\midrule
\multicolumn{5}{l}{\textit{Loss Functions}} \\
\textbf{BCE only (ours)}                & \textbf{63.04} & \textbf{70.36} & 59.19 & 86.71 \\
Contrast only                  &  57.20 & 69.27 &54.39 & \textbf{95.38} \\
BCE + Contrast  & 61.39 & 64.13 & \textbf{60.49} & 68.25 \\
\midrule
\multicolumn{5}{l}{\textit{Pose Encoding}} \\
1D Conv                   & 53.65 & \textbf{67.58} & 52.29 & 95.53 \\
\textbf{2D Conv (ours)}            & \textbf{61.39} & 64.13 & \textbf{60.49} & \textbf{68.25} \\
\bottomrule
\end{tabular}
\end{table}


\subsection{Ablation study and analysis}
To evaluate the robustness of our model, we conduct some ablations using different training choices with a concentration on Accuracy and F1 Scores.

\subsubsection{Effect of different loss function}
As shown in Table~\ref{tab:ablation} demonstrate that using only contrastive loss underperforms when compared to using BCE, indicating that contrastive supervision is not satisfactory enough for this task. While combining both losses with contrastive weight $\lambda=0.5$ achieves 61.39\% accuracy, the result is still below BCE-only performance. The contrastive objective may interfere with classification if the weight is not carefully tuned; using either higher or lower weights results in lower performance, and the embedding space learned through mean pooling may be less discriminative than the [CLS] token representation for this verification task.

\subsubsection{Effect of Pose Encoding}
Table~\ref{tab:ablation} demonstrates that 2D-CNN outperforms other methods in encoding postures. 1D-CNN captures temporal patterns but treats keypoints as a sequence without using their geometric correlations. In contrast, 2D-CNN preserves spatial structure by reshaping each frame as an $n \times 2$ grid, where n is the number of keypoints, allowing the network to learn spatial patterns such as hand configurations and body postures.

\section{Conclusion}
In the work, we present the first video-to-video word presence verification in sign language, where both the sentence and query are video sequences. Our approach proposes using pose sequence in, combining 2D CNN encoding with a Transformer temporal model, achieving 61.66\% accuracy on the \textit{word presence dataset}. 

To the best of our knowledge at the time of this research, no prior work has been done in video-to-video sentence-to-query word spotting. Ablation studies and analysis show that 2D spatial encoding of poses and BCE loss are critical design choices. Our work establishes a strong baseline for this task and demonstrates the effectiveness of pose-based representations for SL understanding.

\bibliography{custom}
\end{document}